# Self Sparse Generative Adversarial Networks

Wenliang Qian[1,2]    Yang Xu[1,2]    Wangmeng Zuo[3]    Hui Li[1,2,4]

[1]Key Lab of  Smart Prevention and Mitigation of Civil Engineering Disaster of the Ministry of Industry and Information Technology

[2]School of Civil Engineering, Harbin Institute of Technology

[3]School of Computer Science and Technology, Harbin Institute of Technology

[4]AI Research Institute, Harbin Institute of Technology

wl.qian@outlook.com    {xyce, wmzuo, lihui}@hit.edu.cn

**Abstract:** *Generative Adversarial Networks (GANs) are an unsupervised generative model that learns data distribution through adversarial training. However, recent experiments indicated that GANs are difficult to train due to the requirement of optimization in the high dimensional parameter space and the zero gradient problem. In this work, we propose a Self Sparse Generative Adversarial Network (Self-Sparse GAN) that reduces the parameter space and alleviates the zero gradient problem. In the Self-Sparse GAN, we design a Self-Adaptive Sparse Transform Module (SASTM) comprising the sparsity decomposition and feature-map recombination, which can be applied on multi-channel feature maps to obtain sparse feature maps. The key idea of Self-Sparse GAN is to add the SASTM following every deconvolution layer in the generator, which can adaptively reduce the parameter space by utilizing the sparsity in multi-channel feature maps. We theoretically prove that the SASTM can not only reduce the search space of the convolution kernel weight of the generator but also alleviate the zero gradient problem by maintaining meaningful features in the Batch Normalization layer and driving the weight of deconvolution layers away from being negative. The experimental results show that our method achieves the best FID scores for image generation compared with WGAN-GP on MNIST, Fashion-MNIST, CIFAR-10, STL-10, mini-ImageNet, CELEBA-HQ, and LSUN bedrooms, and the relative decrease of FID is 4.76% ~ 21.84%.*

## 1 INTRODUCTION

Generative adversarial networks (GANs) [1] are a kind of unsupervised generation model based on game theory, and widely used to learn complex real-world distributions based on deep convolutional layers [2] (e.g. image generation). However, despite its success, training GANs is very unstable, and there may be problems such as gradient disappearance, divergence, and mode collapse [3, 4]. The main reason is that training GANs needs to find a Nash equilibrium for a non-convex problem in a high dimensional continuous space [5]. In addition, it is pointed out that the loss function used in the original GANs [1] causes the zero gradient problem when there is no overlap between the generated data distribution and the real data distribution [6].

The stabilization of GAN training has been investigated by either modifying the network architecture [2, 6-8] or adopting an alternative objective function [9-12]. However, these methods do not reduce the high-dimensional parameter space of the generator. When the task is complex (including more texture details and with high resolution), we often increase the number of convolution kernels to enhance the capability of the generator. Nevertheless, we do not exactly know how many convolution kernels are appropriate, which further increases the parameter space of the generator. Therefore, it is reasonable to speculate that parameter redundancy exists in the generator. If the parameter space of the generator can be reduced, both the performance and training stability of GANs will be further improved.

Motivated by the aforementioned challenges and the sparsity in deep convolution networks [13, 14], we propose a Self-Sparse Generative Adversarial Network (Self-Sparse GAN), with a Self-Adaptive Sparse Transform Module (SASTM) after each deconvolution layer. The SASTM consisting of the sparsity decomposition and feature-map recombination is applied on multi-channel feature maps of the deconvolution layer to obtain sparse feature maps. The channel sparsity coefficients and position sparsity coefficients are obtained by using a two-headed neural network to transform the latent vector in the sparsity decomposition. Then, the sparse multi-channel feature maps are acquired by a superposition of the channel sparsity and position sparsity, which can be obtained by the feature maps multiplying the corresponding sparsity coefficients. The corresponding sparsity coefficients will alleviate the zero gradient problem by maintaining meaningful features in the Batch Normalization (BN) layer and driving the weights of deconvolution layers away from being negative. Meanwhile, the sparse feature maps will free some of the convolution kernels, that is, the weights do not affect the model, thus reducing the parameter space.

**Our contributions.** We propose a novel Self-Sparse GAN, in which the training of generator considers the adaptive sparsity in multi-channel feature maps. We use the SASTM to implement feature maps sparsity adaptively, and theoretically prove that our method not only reduces the search space of the convolution kernel weight but also alleviates the zero gradient problem. We evaluate the performance of proposed Self-Sparse GAN using the MNIST [15], Fashion- MNIST [16], CIFAR-10 [17], STL-10 [18], mini-ImageNet [19], CELEBA-HQ [7], LSUN bedrooms [20] datasets. The experimental results show that our method achieves the best FID scores for image generation compared with WGAN-GP, and the relative decrease of FID is 4.76% ~ 21.84%.

## 2 Related Work

**Generative adversarial network.** GANs [1] can learn the data distribution through the game between the generator and discriminator, and have been widely used in image generation [21], video generation [22], image translation [23], and image inpainting [24].

**Optimization and training frameworks.** With the development of GANs, more and more researchers are committed to settling the training barriers of gradient disappearance, divergence and mode collapse. In the work [5], noise is added to the generated data and real data to increase the support of two distributions and alleviate the problem of gradient disappearance. In the work [10], the least squares loss function is adopted to stabilize the training of the discriminator. WGAN [9] uses the Earth Mover's Distance (EMD) instead of the Jensen-Shannon divergence in the original GAN, which requires the discriminator to satisfy the Lipschitz constraint and can be achieved by weight clipping. Because weight clipping will push weights towards the extremes of the clipping range, WGAN-GP [11] uses the gradient penalty to make the discriminator satisfy the Lipschitz constraint. Another way to enforce the Lipschitz constraint is proposed in [12] by spectral normalization. A series of adaptive methods with the transformation of the latent vector to get additional information are also widely used in GANs. Work [25] uses an adaptive affine transformation to utilize spatial feedback from the discriminator to improve the performance of GANs. Work [26] uses a nonlinear network to transform the latent space to obtain the intermediate latent space, which controls the generator through adaptive instance normalization (AdaIN) in each convolutional layer. In the work [27], a nonlinear network is used to transform the latent vector to obtain the affine transformation parameters of the BN layer to stabilize the GAN training. SPGAN [28] creates a sparse representation vector for each image patch then synthesizes the entire image by multiplying generated sparse representations to a pre-trained dictionary and assembling the resulting patches.

**Sparsity in Convolutional Neural Networks.** Deep convolution networks have made great progress in a wide range of fields, especially for image classification [29]. However, there is a strong correlation between the performance of the network and the network size [30], which also leads to parameter redundancy in deep convolutional networks. The Sparse Convolutional Neural Networks [13, 14, 31] uses sparse decomposition of the convolution kernels to reduce more than 90% parameters, while the drop of accuracy is less than 1% on the ILSVRC2012 dataset. Work [13, 14, 31] proposes $L_0$ norm regularization for neural networks to encourage weights to become exactly zero to speed up training and improve generalization.

## 3 Self-Sparse GAN

Motivated by the aforementioned challenges, we aim to design the generator with a mechanism, which can use fewer feature maps to learn useful representations. Inspired by DANet [32], we first design a two-headed neural network to transform the latent vector to obtain the channel sparsity coefficient and position sparsity coefficient of the multi-channel feature maps. Second, we multiply the multi-channel feature maps by the channel sparse coefficient and position sparse coefficient, respectively. Then, we add the results to get the output of SASTM.

The proposed Self-Sparse GAN adds a SASTM behind each deconvolution layer of the generator. Self-Sparse GAN only modifies the architecture of the generator, and its conceptual diagram is shown in Figure 1.

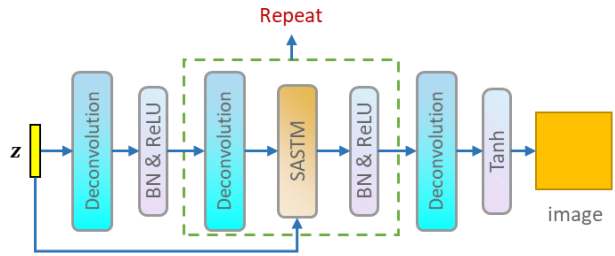

Figure 1: The concept diagram of the modified generator in the Self-Sparse GAN. Deconvolution, BN, ReLU, and Tanh represent the deconvolutional layer, the Batch Normalization layer, rectified linear activation function, and tanh activation function, respectively.

We define the process transforming the planar size of feature map from $H \times W$ to $2H \times 2W$ as a generative stage. For example, when the resolution of the generated image is $128 \times 128$, the hierarchical processes of feature map generation $\mathbf{z} \rightarrow 4 \times 4 \rightarrow 8 \times 8 \rightarrow 16 \times 16 \rightarrow 32 \times 32 \rightarrow 64 \times 64 \rightarrow 128 \times 128$ are defined as different stages in the generator. Stage $t = 3$ refers to $8 \times 8 \rightarrow 16 \times 16$, where $t \in \{1,2,3,4,5,6\}$ and $T = 6$ denotes the total number of stages.

### 3.1. SASTM: Self-Adaptive Sparse Transform Module

SASTM includes the sparsity decomposition and feature-map recombination. The sparsity decomposition consists of Channel Sparsity Module (CSM) and Position Sparsity Module (PSM) to obtain the channel sparsity coefficient and position sparsity coefficient.

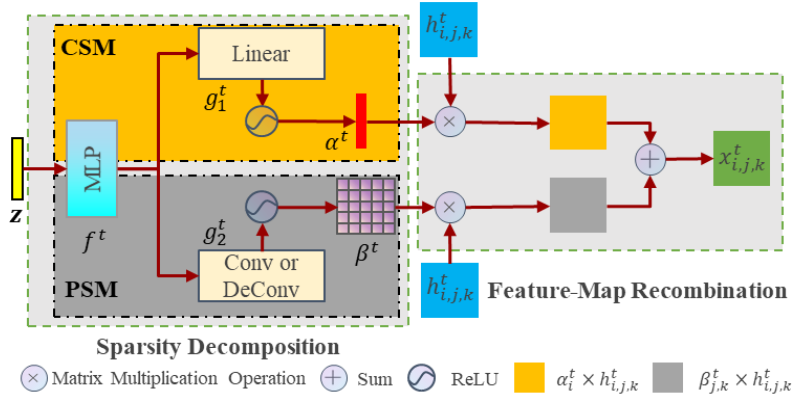

Figure 2: The concept diagram of the Self-Adaptive Sparse Transform Module.

As illustrated in Figure 2, a two-headed neural network will be employed to obtain the corresponding sparsity coefficients. In the two-headed neural network, the underlying shared layers MLP are defined as $f^t$, and the exclusive networks are $g_1^t$ and $g_2^t$, respectively. $\alpha_i^t \geq 0$ and $\beta_{j,k}^t \geq 0$ can be obtained as follows:

$$\boldsymbol{\alpha^t} = ReLU(g_1^t(f^t(\boldsymbol{z})))$$

(1)

$$\boldsymbol{\beta^t} = ReLU(g_2^t(f^t(\boldsymbol{z})))$$

(2)

where Eq.(1) and Eq.(2) represent CSM and PSM, respectively. $\alpha_i^t \in R^{C^t}$ and $\beta_{j,k}^t \in R^{H^t \times W^t}$ are coefficients of the channel sparsity and position sparsity, respectively. When $\alpha_i^t = 0$ and $\beta_{j,k}^t = 0$, the corresponding channel and spatial location will become useless, respectively.

Suppose that the output after the deconvolution layer is $h_{i,j,k}^t \in R^{C^t \times H^t \times W^t}$, where $C^t$, $H^t$ and $W^t$ represent the number of channels, height and width of the feature maps, respectively. The feature-map recombination will be calculated as follow:

$$x_{i,j,k}^t = \alpha_i^t \times h_{i,j,k}^t + \beta_{j,k}^t \times h_{i,j,k}^t$$

(3)

where $x_{i,j,k}^t \in R^{C^t \times H^t \times W^t}$ is sparse feature maps. Therefore, SASTM is the superposition of channel sparsity and position sparsity.

The sparse rate of position sparsity in the $i$-th channel is defined as follows:

$$\xi_i^t = \frac{crad(\{x_{i,j,k}^t, x_{i,j,k}^t = 0\})}{H^t W^t}$$

(4)

where $"crad"$ is used to signify number of elements in the set. If $\xi_i^t \geq 2/3$, the $i$-th channel is regarded to be sparse.

The sparse rate of the channel sparsity is defined as follows:

$$\zeta = \frac{crad(\{\xi_i^t, \xi_i^t \geq 2/3\})}{C^t}$$

(5)

In the BP (Back Propagation) process, the derivatives of loss function with respect to $h_{i,j,k}^t$, $\alpha_i^t$ and $\beta_{j,k}^t$ are calculated as follows:

$$\nabla_{h_{i,j,k}^t} = \nabla_{x_{i,j,k}^t}\left(\frac{\partial x_{i,j,k}^t}{\partial h_{i,j,k}^t}\right) = \nabla_{x_{i,j,k}^t}(\alpha_i^t + \beta_{j,k}^t)$$

$$\nabla_{\alpha_i^t} = \nabla_{x_{i,j,k}^t}\left(\frac{\partial x_{i,j,k}^t}{\partial \alpha_i^t}\right) = \nabla_{x_{i,j,k}^t}h_{i,j,k}^t \qquad (6)$$

$$\nabla_{\beta_{j,k}^t} = \nabla_{x_{i,j,k}^t}\left(\frac{\partial x_{i,j,k}^t}{\partial \beta_{j,k}^t}\right) = \nabla_{x_{i,j,k}^t}h_{i,j,k}^t$$

### 3.2. Mechanism Analysis of the SASTM

To analyze the role of the SASTM, we select the *t*-th generation stage as shown in Figure 3.

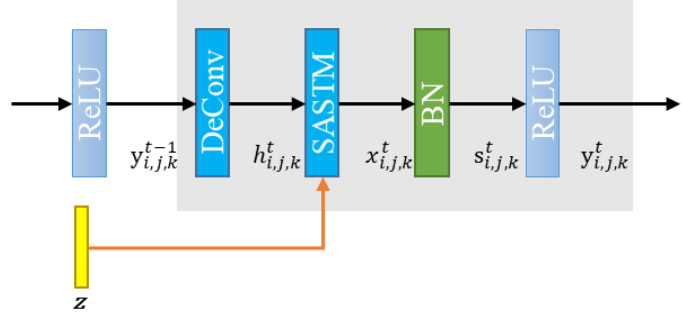

Figure 3: The concept diagram of the *t*-th generation stage. $y_{i,j,k}^{t-1}$, $h_{i,j,k}^t$, $x_{i,j,k}^t$, $s_{i,j,k}^t$ and $y_{i,j,k}^t$ represent the output feature maps of the previous ReLU, deconvolution, SASTM, BN and ReLU layers, respectively, and their dimensions are $C^t \times H^t \times W^t$.

For the convenience of discussion, we assume that dimensions of the input and output feature maps of the deconvolution layer remain same ($C^t \times H^t \times W^t$). Meanwhile, the size of the deconvolution kernel is $1 \times 1$. The feedforward process is expressed as follows:

$$x_{i,j,k}^t = \left(\alpha_i^t \sum_m v_{m,i}^t y_{m,j,k}^{t-1} + \beta_{jk}^t \sum_m v_{m,i}^t y_{m,j,k}^{t-1}\right)$$

$$y_{i,j,k}^t = \max\{\varphi_{bn}^t(x_{i,j,k}^t), 0\}$$

$$(7)$$

where $v_{m,i}^t$ denotes the corresponding deconvolution weight, $y_{m,j,k}^{t-1}$ denotes the element of position $j, k$ in the *m*-th channel, and $\varphi_{bn}^t$ denotes the BN operation.

In the BP process, the derivatives of loss function with respect to $v_{m,i}^t, \alpha_i^t$ and $\beta_{j,k}^t$ are calculated as follows:

$$\nabla_{v_{m,i}^t} = \begin{cases} \nabla_{y_{i,j,k}^t}(\alpha_i^t + \beta_{jk}^t)y_{m,j,k}^{t-1}, & \varphi_{bn}^t(x_{i,j,k}^t) > 0 \\ 0, & \varphi_{bn}^t(x_{i,j,k}^t) \leq 0 \end{cases}$$

$$(8)$$

$$\nabla_{\alpha_i^t} = \begin{cases} \nabla_{y_{i,j,k}^t}\sum_m v_{m,i}^t y_{m,j,k}^{t-1}, & \varphi_{bn}^t(x_{i,j,k}^t) > 0 \\ 0, & \varphi_{bn}^t(x_{i,j,k}^t) \leq 0 \end{cases}$$

$$(9)$$

$$\nabla_{\beta_{jk}^t} = \begin{cases} \nabla_{y_{i,j,k}^t} \sum_m v_{m,i}^t y_{m,j,k}^{t-1} \ , & \varphi_{bn}^t(x_{i,j,k}^t) > 0 \\ 0 \ , & \varphi_{bn}^t(x_{i,j,k}^t) \leq 0 \end{cases}$$

(10)

Intuitively, whether $\alpha_i^t$ and $\beta_{j,k}^t$ are equal to zero or greater than zero will remain unchanged after certain training steps. Therefore, we make the following assumption:

**Hypothesis 1:** When $\alpha_i^t$ and $\beta_{j,k}^t$ have represented the significance of channel and position sparsity, their signs will remain unchanged.

In this section, we prove that the proposed SASTM plays the following three roles:

**1) reducing the search space of convolution parameters in the generator;** In Eq. (7), the dimension of the deconvolution kernel weight $v_{m,i}^t$ is $C^t \times C^t$. Find $\forall i \in A \subseteq \{1, \ldots, C^t\}$, where $\alpha_i^t = 0$, and $\forall j \in \{1, \ldots, H^t\}, \forall k \in \{1, \ldots W^t\}$, where $\beta_{jk}^t = 0$, then $x_{i,j,k}^t = 0$, $\nabla_{v_{m_{i_1}}^t} = 0$ for all times from **hypothesis 1**, which indicates that the $i$-th channel will no longer work in both feedforward and backward processes in training. Consequently, the dimensions of the valid convolutional kernels are $(C^t - |A|) \times C^t$, thus reducing the search space of the convolutional parameters in the generator.

**2) maintaining meaningful features in the BN layer to alleviate the zero gradient problem;** For the convenience of discussion, we do not consider the affine transformation in the BN layer. At the same time, because dividing by the standard deviation in the BN layer will not change the sign, we also ignore the standard deviation in the BN layer for the remainder of the discussion. $x_{i,j,k}^t$ can be divided into two parts $\{x_{i,j,k}^t | x_{i,j,k}^t < 0\}$ and $\{x_{i,j,k}^t | x_{i,j,k}^t \geq 0\}$. When $\{x_{i,j,k}^t | x_{i,j,k}^t < 0\}$ passes through the BN layer, a part of its value will become greater than zero in feedforward and thus mitigate the zero gradient problem in backward. Here, we ignore the part still less than zero. Therefore, in the following discussion, we assume $x_{i,j,k}^t \geq 0$, and denote the positive values as $\tilde{x}_{i,d,e}^t = \{x_{i,d,e}^t, \ x_{i,d,e}^t > 0\}$, $d \in \{1,2, \ldots, D\}$, $e \in \{1,2, \ldots E\}$.

According to the definition of sparse rate of position sparsity, $DE = (1 - \xi_i^t)H^t W^t$. Therefore, the computation of the BN layer can be expressed as

$$s_{i,j,k}^t = x_{i,j,k}^t - \frac{1}{H^t W^t} \sum x_{i,j,k}^t = x_{i,j,k}^t - (1 - \xi_i^t)\frac{1}{DE}\sum \tilde{x}_{i,d,e}^t = x_{i,j,k}^t - (1 - \xi_i^t)\tilde{\mu}_i^t$$

(11)

where $\tilde{\mu}_i^t = \sum \tilde{x}_{i,d,e}^t / DE > 0$ and $s_{i,j,k}^t$ is the value of $x_{i,j,k}^t$ after passing through BN.

The conditional probability of $\{s_{i,j,k}^t < 0 | x_{i,j,k}^t > 0\}$ is $P(\tilde{x}_{i,d,e}^t - \tilde{\mu}_i^t < 0)$. In addition, considering the position sparsity in $x_{i,j,k}^t$, the aforementioned probability is approximated as $P(\tilde{x}_{i,d,e}^t - (1 - \xi_i^t)\tilde{\mu}_i^t < 0)$, where

$$P(\tilde{x}_{i,d,e}^t - \tilde{\mu}_i^t < 0) = P(\tilde{x}_{i,d,e}^t - (1 - \xi_i^t)\tilde{\mu}_i^t < \xi_i \tilde{\mu}_i^t) > P(\tilde{x}_{i,d,e}^t - (1 - \xi_i^t)\tilde{\mu}_i^t < 0)$$

(12)

From Eq. (12), when position sparsity exits in $x_{i,j,k}^t$, the probability of $s_{i,j,k}^t < 0$ will decrease, which increases the probability $y_{ijk}^t > 0$ from Eq. (7). In addition, a larger $\xi_i^t$ will lead to a lower probability of $s_{i,j,k}^t < 0$. Therefore, the gradient in the backpropagation will not disappear. In other words, when $\alpha_i^t$ and $\beta_{jk}^t$ have already determined the sparse channels and spatial locations, SASTM will reduce the likelihood that useful feature information is dropped after passing through the BN layer, thus maintaining meaningful features to alleviate the zero gradient problem.

**3) driving the convolutional weights away from being negative;** From the above derivation, the probability that $\tilde{x}_{i,d,e}^t$ is less than zero after passing through the BN layer can be reduced when the proposed SASTM is implemented into the network, and a larger position sparsity rate $\xi_i^t$ leads to a smaller probability. Therefore, for convenience, we will not consider the BN layer in the discussion below.

When $\nabla_{y_{i,j,k}^t} > 0$ and $x_{i,j,k}^t > 0$, it can be inferred $\nabla_{v_{m,i}^t} > 0$ from Eq. (8), and then $v_{m,i}^t$ will decrease in the gradient descent algorithm. Similarly, according to Eq. (9) and (10), $\nabla_{\alpha_i^t} > 0$ and $\nabla_{\beta_{jk}^t} > 0$ can be inferred, and thus $\alpha_i^t$ and $\beta_{jk}^t$ will decrease. However, from **Hypothesis 1**, $\alpha_i^t$ and $\beta_{jk}^t$ will not be less than zero. According to Eq.(8), $\nabla_{v_{m,i}^t}$ is obtained by the factors $\alpha_i^t$ and $\beta_{jk}^t$. Therefore, the decrease of $v_{m,i}^t$ is suppressed.

When $\nabla_{y_{i,j,k}^t} < 0$ and $x_{i,j,k}^t > 0$, it can be inferred $\nabla_{v_{m,i}^t} < 0$ from Eq. (8), and then $v_{m,i}^t$ will increase. Similarly, according to Eq. (9) and (10), $\nabla_{\alpha_i^t} < 0$ and $\nabla_{\beta_{jk}^t} < 0$ can be inferred, and thus $\alpha_i^t$ and $\beta_{jk}^t$ will increase. Therefore, the increase of $v_{m,i}^t$ is promoted.

Therefore, the proposed SASTM enables to drive convolutional weights away from being negative. This phenomenon has been similarly reported as the Channel Scaling layer [33].

## 4 Experiments

### 4.1. Baseline Model: WGAN-GP

WGAN-GP [11] has good theoretical and stability properties in practice, and a zero-centered gradient penalty further enhances the convergency [3]. Therefore, WGAN with a zero-centered gradient penalty is adopted as the baseline for comparison with our method, and the objective function is as follows:

$$\min_G \max_D V(G, D) = \mathbb{E}_{\mathbf{x} \sim P_r}[D(\mathbf{x})] - \mathbb{E}_{\mathbf{z} \sim P_z}[D(G(\mathbf{z}))] - \lambda \cdot \mathbb{E}_{\hat{\mathbf{x}} \sim P_{\hat{\mathbf{x}}}}[\|\nabla_{\hat{\mathbf{x}}} D(\hat{\mathbf{x}})\|_2^2] \tag{13}$$

where $\mathbf{x}$, $\mathbf{z}$, $\lambda$ and $\hat{\mathbf{x}}$ represent the real data, latent vector, gradient penalty coefficient and random samples with sampling uniformly along straight lines between pairs of real data and fake data [11].

### 4.2. Experimental Settings

We test the proposed Self-Sparse GAN using a DCGAN-like [2] network architecture on seven datasets: 1) MNIST: 60K grayscale images; 2) Fashion-MNIST: 60K grayscale images; 3) CIFAR-10: 60K RGB images; 4) STL-10: 100K RGB images; 5) mini-ImageNet: 60K RGB images; 6) CELEBA-HQ: 30K RGB images; 7) LSUN bedrooms: 3M RGB images. Details are referred to Appendix 1 in the supplementary materials. The investigated resolutions of generated images are listed in Table 1.

| Datasets | Resolution of generated images |
|:---:|:---:|
| MNIST | 32×32, 128×128 |
| Fashion-MNIST | 64×64, 128×128 |
| CIFAR-10 | 128×128×3 |
| STL-10 | 64×64×3 |
| mini-ImageNet | 32×32×3, 64×64×3 |
| CELEBA-HQ | 64×64×3, 128×128×3 |
| LSUN bedrooms | 64×64×3, 128×128×3 |

Table 1: The investigated resolutions of generated images.

We use the Adam [34] optimizer and set the learning rates of generator and discriminator as 0.0001 and 0.0003 on all datasets as suggested in [35]. Because multiple discriminator steps per generator step can help the GAN training in WGAN-GP, we set two discriminator steps per generator step for 100K generator steps. We set $\beta_1 = 0.5, \beta_2 = 0.999$ for MNIST, Fashion-MNIST, CIFAR-10, STL-10 and mini-ImageNet, and $\beta_1 = 0.0, \beta_2 = 0.9$ for CELEBA-HQ and LSUN bedrooms.

For the evaluation of model sampling quality, we use Frechet Inception Distance (FID) [35] as the evaluation metric, which can measure the distance between the real and generated data distributions. A smaller FID indicates better qualities of the generated images. The FID is calculated as:

$$FID(x,g) = \left\| \mu_x - \mu_g \right\|_2^2 + Tr(\Sigma_x + \Sigma_g - 2(\Sigma_x \Sigma_g)^{1/2}) \tag{14}$$

where $\mu$ and $\Sigma$ denote mean and covariance, respectively, and $x$ and $g$ denote the real and generated data, respectively. To obtain training curves quickly, the FID is evaluated every 500 generator steps using the 5K samples.

### 4.3. Results

**Comparison with WGAN-GP:** Figure 4 shows the FID curves on MNIST and CIFAR-10, and those of other datasets are plotted in the Appendix 2 of the supplementary materials. Figure 4 indicates that our method converges faster in the same FID level. Table 2 shows the mean and standard deviation of the best FIDs on all datasets. Experimental results show that our method reduces FIDs on all datasets and the relative decrease of FID is 4.76% ~ 21.84%. Although the Self-Sparse GAN does not significantly exceed the baseline on CELEBA-HQ with the resolutions of 64×64×3, the relative decrease of FID is still close to 5%. Meanwhile, these results demonstrate that our method can both improve the generation quality of grayscale and RGB images. In addition, the relative improvement of model performance increases with the resolution of generated images from $64 \times 64 \times 3$ to $128 \times 128 \times 3$ on CELEBA-HQ and LSUN bedrooms.

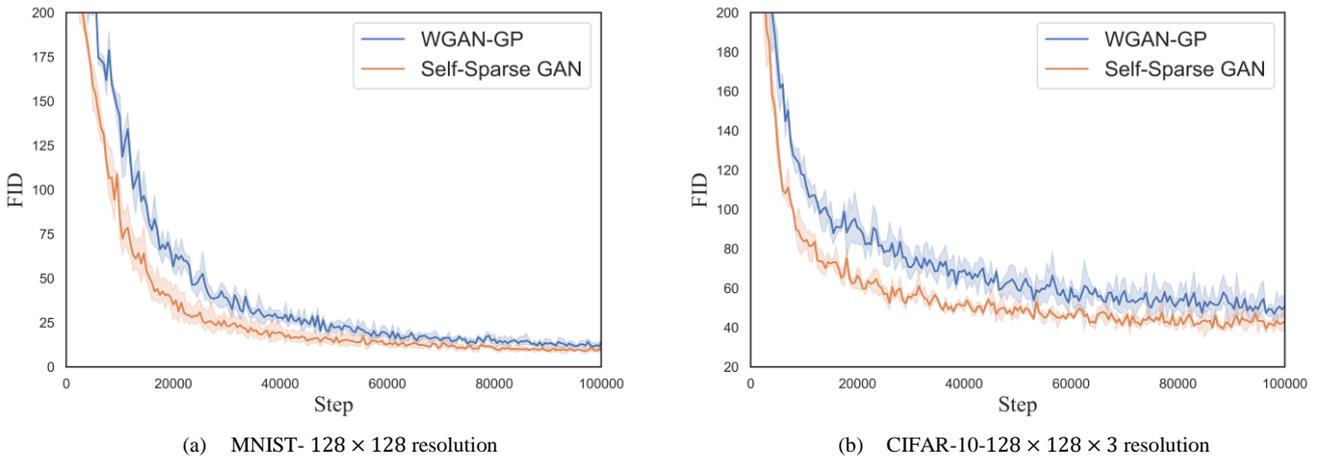

(a)   MNIST- 128 × 128 resolution                          (b)   CIFAR-10-128 × 128 × 3 resolution

Figure 4: FID training curves on MNIST and CIFAR10, depicting the mean performance of three random trainings with a 95% confidence interval.

| Datasets | Resolution | Model | FID | Datasets | Resolution | Model | FID |
|---|---|---|---|---|---|---|---|
| MNIST (Grayscale) | $32\times32\times1$ | WGAN-GP | $7.43 \pm 0.28$ | mini-ImageNet (RGB) | $32\times32\times3$ | WGAN-GP | $33.16 \pm 0.02$ |
| | | Ours | $\mathbf{6.26 \pm 0.64}$ | | | Ours | $\mathbf{28.88 \pm 0.37}$ |
| | $128\times128\times1$ | WGAN-GP | $10.42 \pm 0.86$ | | $64\times64\times3$ | WGAN-GP | $58.81 \pm 3.28$ |
| | | Ours | $\mathbf{8.32 \pm 1.03}$ | | | Ours | $\mathbf{54.78 \pm 0.28}$ |
| Fashion-MNIST (Grayscale) | $64\times64\times1$ | WGAN-GP | $20.37 \pm 0.87$ | CELEBA-HQ (RGB) | $64\times64\times3$ | WGAN-GP | $15.95 \pm 0.44$ |
| | | Ours | $\mathbf{15.92 \pm 1.10}$ | | | Ours | $\mathbf{15.19 \pm 0.18}$ |
| | $128\times128\times1$ | WGAN-GP | $20.41 \pm 0.70$ | | $128\times128\times3$ | WGAN-GP | $32.40 \pm 2.03$ |
| | | Ours | $\mathbf{17.67 \pm 0.87}$ | | | Ours | $\mathbf{27.72 \pm 1.54}$ |
| CIFAR-10 (RGB) | $128\times128\times3$ | WGAN-GP | $43.77 \pm 2.10$ | LSUN bedrooms (RGB) | $64\times64\times3$ | WGAN-GP | $59.12 \pm 0.95$ |
| | | Ours | $\mathbf{36.69 \pm 1.53}$ | | | Ours | $\mathbf{55.06 \pm 2.00}$ |
| STL-10 (RGB) | $64\times64\times3$ | WGAN-GP | $63.88 \pm 1.33$ | | $128\times128\times3$ | WGAN-GP | $102.16 \pm 0.85$ |
| | | Ours | $\mathbf{56.23 \pm 1.38}$ | | | Ours | $\mathbf{84.78 \pm 2.89}$ |

Table 2:  Comparison of FIDs between our proposed Self-Sparse GAN and the baseline WGAN-GP. The mean and standard deviation of the FID are calculated through three individual training with different random seeds.

On multiple datasets, our reported FIDs for WGAN-GP may be smaller than other literature. For example, On the STL-10 dataset, the reported FID for the WGAN-GP is **63.88**, which is smaller than **55.1** reported by [36]. The main reason is that we used 5K samples in a resolution of $64 \times 64 \times 3$ to calculate FID rather than using 50K samples in a resolution of $48 \times 48 \times 3$ in [36]. Table 3 shows the new FIDs by 50K samples, which shows the FID is better than 55.1 even with $64 \times 64 \times 3$ resolution. Therefore, this paper doesn't lower the baseline.

| Samples | WGAN-GP | Self-Sparse GAN |
|---|---|---|
| 5K | 63.88 | **56.26** |
| 50K | 54.23 | **46.85** |

Table 3: The FIDs in STL-10 with resolution of $64 \times 64 \times 3$ by 50K samples

### 4.4. Ablations

To investigate the effects of CSM and PSM in the proposed SASTM, we perform ablation studies on Fashion-MNIST and STL-10. "Without PSM" represents using CSM only and $\beta_{j,k}^t = 0$. Similarly, "without CSM" represents using PSM only and $\alpha_i^t = 0$.

As shown in Table 4, the model performance has a significant improvement on Fashion-MNIST and STL-10 when both CHM and PSM are applied. Since the position sparsity coefficient $\boldsymbol{\beta^t}$ is shared by all channels, it is difficult to represent the pixel-wise sparsity among different channels without $\boldsymbol{\alpha^t}$. Therefore, using only PSM may not function well. On the other hand, when only CSM is used, the model may lack generation power. Figure 5 also shows that using only CSM on the Fashion-MNIST dataset causes the multi-channel feature maps too sparse, which will suppress the model performance.

| Datasets | Resolution | Method | FID |
|---|---|---|---|
| Fashion-MNIST | 128×128×1 | WGAN-GP | 20.41 ±0.70 |
| | | Ours | **17.67 ±0.87** |
| | | Without PSM | 21.51 ±0.29 |
| | | Without CSM | 163.71 ±3.99 |
| STL-10 | 64×64×3 | WGAN-GP | 63.88 ± 1.33 |
| | | Ours | **56.23 ± 1.38** |
| | | Without PSM | 60.85 ±0.99 |
| | | Without CSM | 64.87 ±1.71 |

Table 4: Comparisons of FIDs in ablation studies on Fashion-MNIST and STL-10.

| Datasets | $\beta_1, \beta_2$ | Method | FID |
|---|---|---|---|
| STL-10 | 0, 0.9 | WGAN-GP | 63.88 ±1.33 |
| | 0, 0.9 | Ours | **56.23 ±1.38** |
| | 0.5,0.999 | WGAN-GP | 67.13 ±0.77 |
| | 0.5,0.999 | Ours | **56.51 ±1.62** |

Table 5: Comparisons of FID in the robustness experiments on STL-10 with different Adam hyperparameter settings.

| Datasets | Arch | Method | FID |
|---|---|---|---|
| STL-10 | DCGAN | WGAN-GP | 63.88 ± 1.33 |
| | | Ours | **56.23 ± 1.38** |
| | ResNet | WGAN-GP | 65.16 ±5.96 |
| | | Ours | **60.31 ±4.29** |

Table 6: Comparisons of FID in the robustness experiments on STL-10 with different network architectures.

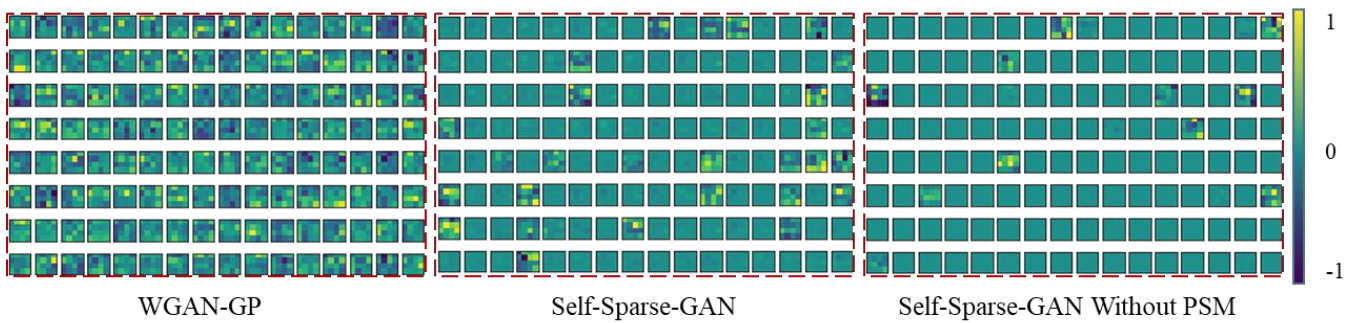

Figure 5: Ablation study. Refer to Section 4.4 for details. Using only CHM causes too sparse multi-channel feature maps.

**Robustness to Hyperparameters of Adam Training.** GANs are very sensitive to hyperparameters of the optimizer. Therefore, we evaluate different hyperparameter settings to validate the robustness of our method. We test two popular settings of $(\beta_1, \beta_2)$ in Adam: (0, 0.9) and (0.5, 0.999). Table 5 compares the mean and standard deviation of FID scores on STL-10. It suggests that the proposed Self-Sparse GAN consistently improves model performance.

**Robustness to Network Architectures.** To further test the robustness of the proposed Self-Sparse GAN to different network architectures, we use two common network architectures from DCGAN and ResNet on STL-10. Details are referred to Appendix 3 in the supplementary materials. Table 6 compares the FID scores using different network architectures on STL-10, which shows that our method is robust to both DCGAN and ResNet network architectures.

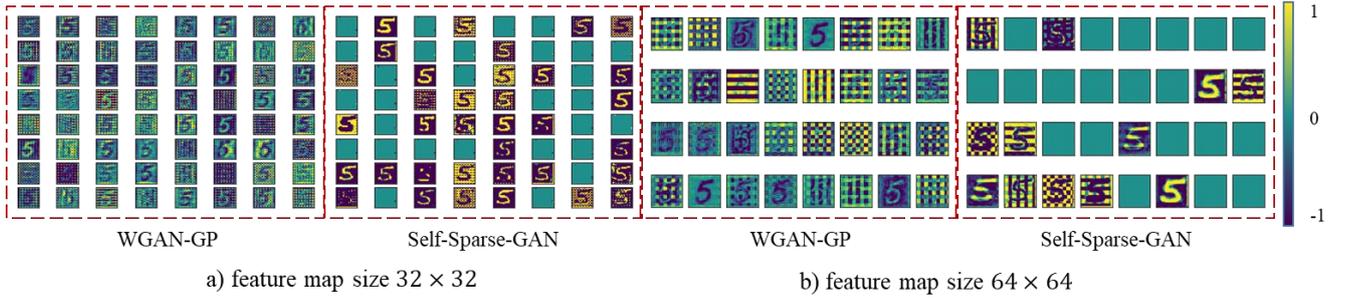

a) feature map size $32 \times 32$         b) feature map size $64 \times 64$

Figure 6: Visualization of the feature map of the output of SASTM. It can be observed that Self-Sparse GAN learns to pick useful convolutional kernels instead of using all convolutional kernels for image generation. In a), we can observe that some sparse feature maps have regular feature points, which means that the PSM is working.

### 4.5. Visualization of SASTM Features

To illustrate the function of SASTM, we visualize the multi-channel feature maps of each deconvolution layer in the generator on MNIST with a resolution of 128×128. Figure 6 shows multi-channel feature maps in two representative levels of 64 ×64 and 32 ×32, and results in other spatial sizes are shown in Appendix 4 of the supplementary materials. Results show that the proposed Self-Sparse GAN learns to pick useful sparse convolutional kernels instead of using all kernels greedily. It also proves that our method can obtain sparse multi-channel feature maps, thus reducing network parameters.

**Validation of Hypothesis 1:** We can verify this hypothesis by visualizing feature maps under different training steps on MNIST, as shown in Figure 7. The results illustrate that the sign of $\alpha_i^t$ and $\beta_{j,k}^t$ will remain unchanged after 5000 generator steps.

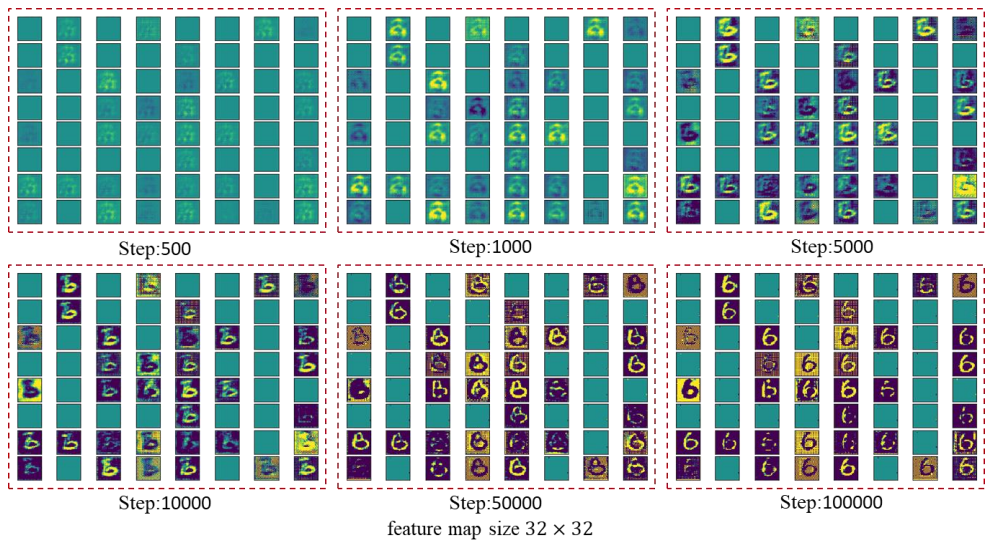

feature map size $32 \times 32$

Figure 7 Validation of Hypothesis 1 on MNIST with the resolution of 128×128

## 4.6. Investigation of Relationship between Sparsity and FID

In Section 3.2, we have proved that the proposed SASTM will alleviate the zero gradient problem and thus improve the model performance. To analyze the relationship between model sparsity and FID quantitatively, we define the average position sparsity rate $\bar{\xi}$ of the generators as

$$\bar{\xi} = \frac{1}{T} \sum_t \left( \frac{1}{C^t} \sum_i \xi_i^t \right), t \in \{1,2,\dots T\}, i \in \{1,2,\dots C^t\} \tag{15}$$

We select the same network architecture to calculate the corresponding average position sparsity rate according to Eq.(15), as shown in Table 7. From the data, it indicates that a larger average position sparsity rate may lead to a greater improvement in FID except on the mini-ImageNet with $64 \times 64$ resolution and CIFAR-10 with $128 \times 128$ resolution, which will be further investigated in the future study.

The Pearson's coefficient is used to measure the correlation between the average position sparsity rate and FID as

$$\rho(\bar{\xi}, \boldsymbol{\eta}) = \frac{E[(\bar{\xi} - \mu_{\bar{\xi}})(\boldsymbol{\eta} - \mu_{\boldsymbol{\eta}})]}{\sigma_{\bar{\xi}} \sigma_{\eta}} \tag{16}$$

where $\bar{\xi}$ and $\boldsymbol{\eta}$ denote the average position sparsity rate and the relative decrease of FID, respectively. A positive correlation between the average position sparsity rate and FID is found. When the resolution of the generated image is $64 \times 64$, Pearson's correlation coefficient is 0.62. When the resolution of the generated image is $128 \times 128$, Pearson's correlation coefficient is 0.79. Meanwhile, with the increase of resolution, the Pearson's correlation coefficient will increase.

| Resolution | Datasets | Average Position Sparsity Rate | FID Reduction (%) | Resolution | Datasets | Average Position Sparsity Rate | FID Reduction (%) |
|---|---|---|---|---|---|---|---|
| 64×64 | STL-10 | 0.44 | 11.97 | 128×128 | MNIST | 0.53 | 20.21 |
| | mini-ImageNet | 0.47 | 6.83 | | CIFAR-10 | 0.47 | 16.17 |
| | CELEBA-HQ | 0.26 | 4.76 | | CELEBA-HQ | 0.21 | 14.43 |
| | LSUN bedrooms | 0.39 | 6.86 | | LSUN bedrooms | 0.31 | 17.02 |

Table 7: Relationship between sparseness and FID with the same network architecture.

## 5 Conclusions

In this study, a Self Sparse Generative Adversarial Network (Self-Sparse GAN) is proposed for the unsupervised image generation task. By exploiting channel sparsity and position sparsity in multi-channel feature maps, Self-Sparse GAN stabilizes the training process and improves the model performance by: (1) reducing the search space of convolution parameters in the generator; (2) maintaining meaningful features in the BN layer to alleviate the zero gradient problem; (3) driving the convolutional weights away from being negative. We demonstrate the proposed method on seven image datasets. Experimental results show that our approach can obtain better FIDs on all the seven datasets compared with WGAN-GP, and is robust to both training hyperparameters and network architectures. Besides, a positive correlation between sparsity and FID further validates that the proposed sparsity module enhances the image generation power of the model.

## Acknowledgment

Financial support for this study was provided by the National Natural Science Foundation of China [No. 51638007, 51921006, U1711265 and 52008138], National Key R&D Program of China [No. 2018YFC0705605 and 2019YFC1511102], China Post-doctoral Science Foundation [No. BX20190102 and 2019M661286], and Heilongjiang Post-doctoral General Funding [No. LBH-Z19064].